\documentclass[11pt]{article}

\usepackage[final]{automl}
\usepackage{microtype} 
\usepackage{booktabs}  
\usepackage{url}  
\usepackage{csquotes}
\usepackage{natbib}

\usepackage[utf8]{inputenc}
\usepackage{url}

\usepackage{amsmath}
\usepackage{bbm}
\usepackage{bm}
\usepackage{cancel}
\usepackage{mathtools}

\usepackage{ascmac}  
\usepackage{algorithm}
\usepackage{algpseudocode}

\usepackage{fancyhdr}
\usepackage{tcolorbox}
\usepackage{tikz}

\usepackage{booktabs}
\usepackage{comment}
\usepackage{lastpage}
\usepackage{listings}
\usepackage{multibib}
\usepackage{multirow}
\usepackage[super]{nth}

\tcbuselibrary{breakable, skins, theorems}

\allowdisplaybreaks

\algtext*{EndFor}
\algtext*{EndWhile}
\algtext*{EndIf}
\algtext*{EndProcedure}
\algtext*{EndFunction}
\algnewcommand{\LineComment}[1]{\State \(\triangleright\) #1}

\newcites{appx}{References}

\hypersetup{%
  pdfauthor={Shuhei Watanabe}, 
  pdftitle={Derivation of Frank Log EI},
  pdfsubject={},
  pdfkeywords={AutoML, LaTeX, style}
}

\newif\ifunderreview
\newif\ifsubmission
\newif\ifappendix

\underreviewfalse

\submissiontrue

\appendixfalse

\ifsubmission
\newcommand{\todo}[1]{}
\newcommand{\replace}[2]{}
\newcommand{\sw}[1]{}

\else
\newcommand{\todo}[1]{\textbf{\textcolor{red}{[TODO: #1]}}}

\newcommand{\replace}[2]{\textbf{\textcolor{red}{[del: \cancel{#1}]}}\textbf{\textcolor{blue}{[new: #2]}}}

\newcommand{\sw}[1]{\textbf{\textcolor{cyan}{[SW: #1]}}}

\usepackage[inline]{showlabels}

\fi

\makeatletter
\newcommand{\customlabel}[2]{%
   \protected@write \@auxout {}{\string \newlabel {#1}{{#2}{\thepage}{#2}{#1}{}} }%
   \hypertarget{#1}{}
}
\makeatother

\newcommand{\xv}{\boldsymbol{x}}
\newcommand{\D}{\mathcal{D}}
\newcommand{\expf}[1]{\exp\biggl(#1\biggr)}

\renewcommand{\eqref}[1]{Eq.~(\ref{#1})}

\DeclareFixedFont{\ttb}{T1}{txtt}{bx}{n}{12} 
\DeclareFixedFont{\ttm}{T1}{txtt}{m}{n}{12}  

\definecolor{deepblue}{rgb}{0,0,0.5}
\definecolor{deepred}{rgb}{0.6,0,0}
\definecolor{deepgreen}{rgb}{0,0.5,0}

\begin{document}

\title{Derivation of Closed Form of Expected Improvement for Gaussian Process Trained on Log-Transformed Objective}

\author{
  Shuhei Watanabe \\
  Preferred Networks Inc. \\
  \texttt{shuheiwatanabe@preferred.jp}
}

\maketitle

\begin{abstract}
  Expected Improvement (EI) is arguably the most widely used acquisition function in Bayesian optimization.
  However, it is often challenging to enhance the performance with EI due to its sensitivity to numerical precision.
  Previously, \cite{hutter2009experimental} tackled this problem by using Gaussian process trained on the log-transformed objective function and it was reported that this trick improves the predictive accuracy of GP, leading to substantially better performance.
  Although \cite{hutter2009experimental} offered the closed form of their EI, its intermediate derivation has not been provided so far.
  In this paper, we give a friendly derivation of their proposition.
\end{abstract}

\section{Preliminaries}
In this paper, we consistently consider the maximization problem and use the following notations:
\begin{itemize}
  \item $\xv \in \mathcal{X}$, an input vector defined on the search space $\mathcal{X} \subseteq \mathbb{R}^D$,
  \item $y \in \mathbb{R}$, an objective value, which is better when it is larger,
  \item $\D \coloneqq \{(\xv_n, y_n)\}_{n=1}^N$, a set of observations where $y_n$ is the objective value given an input vector $\xv_n$,
  \item $[N] \coloneqq \{1,\dots,N\}$, a set of integers from $1$ to $N$,
  \item $\Phi: \mathbb{R} \rightarrow [0,1]$, the standard normal distribution function,
  \item $\phi: \mathbb{R} \rightarrow \mathbb{R}_+$, the probability density function of $\Phi$, i.e. $\frac{d\Phi}{du} = \phi(u) = \frac{1}{\sqrt{2\pi}}\exp(-\frac{u^2}{2})$,
  \item $\mu(\xv)$, the mean at $\xv$ predicted by Gaussian process (GP) trained on $\D$, and
  \item $\sigma(\xv)$, the standard deviation at $\xv$ predicted by GP trained on $\D$.
\end{itemize}
Training GP means that we can predict the following:
\begin{equation}
\begin{aligned}
  y \sim \mathcal{N}(\mu(\xv), \sigma(\xv)^2),
\end{aligned}
\end{equation}
and we use the probability density function of this distribution as the posterior $p(y | \xv, \D)$.
Note that since this paper focuses only on the derivation of the closed form provided by \cite{hutter2009experimental,hutter2011sequential}, we defer the details of Bayesian optimization to other works such as \cite{brochu2010tutorial,shahriari2016taking,garnett2022bayesian}.

\section{Derivation of Closed Form Solutions}
In this section, we show how to derive the closed-form solution provided by \cite{hutter2009experimental}.
To do so, we first begin with deriving the closed-form solution of expected improvement (EI)~\citep{jones1998efficient}.

\subsection{Expected Improvement}
First of all, EI is the following acquisition function:
\begin{equation}
\begin{aligned}
  \alpha_{\mathrm{EI}}(\xv | \D) &= \int_{-\infty}^{y^\star} (y^\star - y) p(y | \xv, \D) dy
\end{aligned}
\end{equation}
where $y^\star \in \mathbb{R}$ is usually $\max_{n \in [N]} y_n$.
With GP trained on $\D$, we can transform EI as follows:
\begin{equation}
  \begin{aligned}
    \alpha_{\mathrm{EI}}(\xv | \D) &= \int_{-\infty}^{y^\star} (y^\star - y) p(y | \xv, \D) dy \\
    &= \int_{-\infty}^{y^\star} (y^\star - y) \frac{1}{\sqrt{2\pi \sigma^2}}\expf{-\frac{(y - \mu)^2}{2\sigma^2}} dy ~(\mathrm{Def.}~\mu \coloneqq \mu(\xv), \sigma \coloneqq \sigma(\xv))\\
    &= \int_{-\infty}^{z} (y^\star - \mu - u\sigma) \frac{1}{\sqrt{2\pi \sigma^2}}\expf{
      -\frac{u^2}{2}
    } (\sigma du) ~\biggl(
      \mathrm{Def.}~u \coloneqq \frac{y - \mu}{\sigma}, \frac{du}{dy} = \frac{1}{\sigma}, z \coloneqq \frac{y^\star - \mu}{\sigma}
    \biggr)\\
    &= \int_{-\infty}^{z} (y^\star - \mu - u\sigma)\phi(u)du \\
  \end{aligned}
  \label{eq:basic-ei-transformation}
\end{equation}
Since $\Phi(z) = \int_{-\infty}^{z} \phi(u)du$ by definition, and $y^\star$ and $\mu$ does not depend on $u$, the first and second terms can be computed as $\int_{-\infty}^{z} (y^\star - \mu)\phi(u)du = (y^\star - \mu) \Phi(z) = z\sigma\Phi(z)$.
The last term can be computed as follows:
\begin{equation}
\begin{aligned}
  \int_{-\infty}^{z} u\phi(u)du &= \int_{-\infty}^{z} \frac{u}{\sqrt{2\pi}}\expf{-\frac{u^2}{2}}du \\
  &= -\int_{-\infty}^{-\frac{z^2}{2}} \frac{1}{\sqrt{2\pi}}\exp v \underbrace{dv}_{=-udu}~\biggl(\mathrm{Def.}~v\coloneqq -\frac{u^2}{2}, \frac{dv}{du} = -u\biggr) \\
  &= -\frac{1}{\sqrt{2\pi}}[\exp v]_{-\infty}^{-\frac{z^2}{2}} 
  = -\frac{1}{\sqrt{2\pi}}\expf{-\frac{z^2}{2}} = -\phi(z).
\end{aligned}
\end{equation}
Therefore, the closed-form solution of EI is:
\begin{equation}
\begin{aligned}
  \alpha_{\mathrm{EI}}(\xv | \D) = z\sigma\Phi(z) -\sigma (-\phi(z)) = z\sigma\Phi(z) +\sigma \phi(z).
\end{aligned}
\end{equation}

\subsection{Expected Improvement Proposed by \cite{hutter2009experimental}}

We now consider EI for the log-transformed objective, i.e. Eq.~(5) by \cite{hutter2009experimental}, where $l = \log y$ is the log-transformed objective and GP is trained using $l_n \coloneqq \log y_n$ instead of $y_n$, meaning that $l \sim \mathcal{N}(\mu(\xv), \sigma(\xv)^2)$:
\begin{equation}
\begin{aligned}
  \alpha_{\log\mathrm{EI}}(\xv | \D) &= \int_{-\infty}^{\log y^\star} (y^\star - \exp l) p(l | \xv, \D) dl. \\
\end{aligned}
\end{equation}
By defining $z = \frac{\log y^\star - \mu}{\sigma}$ and $u = \frac{l - \mu}{\sigma}$, we obtain the following in a similar way as \eqref{eq:basic-ei-transformation}:
\begin{equation}
\begin{aligned}
  \alpha_{\log\mathrm{EI}}(\xv | \D) 
  = \int_{-\infty}^{z} (y^\star - \exp(\mu + u\sigma))\phi(u)du = y^\star \Phi(z) - \exp \mu \int_{-\infty}^{z} \exp(u\sigma)\phi(u)du. \\
\end{aligned}
\label{eq:frank-ei-basic-transformation}
\end{equation}
The second term can be transformed by completing the square as follows:
\begin{equation}
\begin{aligned}
  \int_{-\infty}^{z} \exp(u \sigma)\phi(u) du &= \int_{-\infty}^{z} \frac{1}{\sqrt{2\pi}}\expf{
    -\frac{u^2}{2} + u \sigma
  } du \\
  &= \int_{-\infty}^{z} \frac{1}{\sqrt{2\pi}}\expf{
    -\frac{(u - \sigma)^2}{2} + \frac{\sigma^2}{2}
  } du \\
  &= \exp{
    \frac{\sigma^2}{2}
  } \int_{-\infty}^{z - \sigma} \frac{1}{\sqrt{2\pi}}\expf{
    -\frac{v^2}{2}
  } dv~\biggl(
    \mathrm{Def.}~v \coloneqq u - \sigma, \frac{dv}{du} = 1
  \biggr) \\
  &= \exp \frac{\sigma^2}{2} \Phi(z - \sigma).
\end{aligned}
\label{eq:second-term-of-frank-ei}
\end{equation}
Combining \eqref{eq:frank-ei-basic-transformation} and \eqref{eq:second-term-of-frank-ei}, we yield:
\begin{equation}
\begin{aligned}
  \alpha_{\log\mathrm{EI}}(\xv | \D) &= y^\star \Phi(z) - \exp \mu \exp \frac{\sigma^2}{2}
  \Phi(z - \sigma) = y^\star \Phi(z) - \expf{
    \mu + \frac{\sigma^2}{2}
  } \Phi(z - \sigma) \\
\end{aligned}
\end{equation}
where $z = \frac{\log y^\star - \mu}{\sigma}$.
The final result is identical to the closed-form solution, i.e. Eq.~(6) provided by \cite{hutter2009experimental}.

\section{Related Work}
The acquisition function discussed in this paper is primarily used by SMAC3~\citep{lindauer2022smac3}.
As mentioned by \cite{hutter2009experimental}, the logarithmic transformation increases the predictive accuracy of GP, leading to better performance.
As this acquisition function involves logarithmic transformation, this function is conventionally called \texttt{logEI}.
However, we need to emphasize that this acquisition function is very different from the acquisition function discussed by \cite{ament2024unexpected} where the authors literally take the logarithm of EI, i.e. $\log \alpha_{\mathrm{EI}}(\xv |\D)$.
For example, \texttt{logEI} defined in BoTorch~\citep{balandat2020botorch} and \texttt{GPSampler} of Optuna~\citep{akiba2019optuna} is $\log \alpha_{\mathrm{EI}}(\xv |\D)$.
Note that \texttt{TPESampler}, the main Bayesian optimization algorithm used in Optuna, also uses $\log \alpha_{\mathrm{EI}}(\xv |\D)$ for the numerical stability.
We defer the details of the TPE algorithm to \cite{bergstra2011algorithms,watanabe2023tree}.

\section{Conclusion}

In this paper, we showed the detailed derivation of the closed-form solution of EI proposed by \cite{hutter2009experimental}.
We hope that our derivation helps practitioners verify their implementations and works as a future reference if anyone would like to further enhance this acquisition function.

\newpage

\bibliographystyle{apalike}
\bibliography{ref}

\end{document}


\ifappendix

\else

\customlabel{data1}{1}
\customlabel{data2}{2}

\renewcommand\thefigure{\arabic{figure}}
\setcounter{figure}{1}  
\renewcommand\thetable{\arabic{table}}
\setcounter{table}{2}  
\renewcommand\thealgorithm{\arabic{algorithm}}
\setcounter{algorithm}{3}  
\renewcommand\theequation{\arabic{equation}}
\setcounter{equation}{4}  

\renewcommand{\thesection}{\Alph{section}} 
\renewcommand{\thesubsection}{\thesection.\arabic{subsection}}
\renewcommand{\thesubsubsection}{\thesection.\arabic{subsection}.\arabic{subsubsection}}

\fi

\section*{Appendix}
Random Search~\citeappx{bergstra2012random}.

\bibliographystyleappx{splncs04}  
\bibliographyappx{ref}